\documentclass[runningheads]{llncs}
\usepackage[T1]{fontenc}
\usepackage{graphicx}
\usepackage[hidelinks]{hyperref}
\usepackage{color}

\usepackage{amsmath,amssymb,mathtools}
\usepackage[dvipsnames]{xcolor}
\usepackage[capitalise]{cleveref}
\usepackage{subcaption}
\usepackage{siunitx}
\usepackage{booktabs}
\usepackage[stable]{footmisc}

\usepackage{tikz}
\usetikzlibrary{bayesnet}

\usepackage{algorithmic}
\usepackage{algorithm}

\newcommand{\bR}{\mathbb{R}}
\newcommand{\mbi}{\mathbf{i}}

\newcommand{\ten}[1]{\boldsymbol{\mathcal{\uppercase{#1}}}}
\newcommand{\mat}[1]{\boldsymbol{\uppercase{#1}}}
\newcommand{\ve}[1]{\boldsymbol{#1}}

\newcommand{\tr}{\mathrm{tr}}

\newcommand{\diag}{\mathrm{diag}}
\newcommand{\kl}[2]{\mathbb{D}_{KL}( #1 \lVert #2 )}
\newcommand{\E}{\mathbb{E}}
\newcommand{\ind}{\boldsymbol{\mathrm{i}}}

\newcommand{\pg}{P\'{o}lya-Gamma }

\newcommand{\eg}{\emph{e.g.}}

\newcommand{\red}[1]{{\color{red} #1}}
\newcommand{\blue}[1]{{\color{blue} #1}}

\begin{document}
%
% \title{Contribution Title\thanks{Supported by organization x.}}
\title{Scalable Bayesian Tensor Ring Factorization for Multiway Data Analysis}
\author{Zerui Tao\inst{1,2} \and
Toshihisa Tanaka\inst{1,2} \and
Qibin Zhao\inst{2,1}\thanks{Corresponding author}}
% \author{Zerui Tao\inst{1}\orcidID{} \and
% Toshihisa Tanaka\inst{2,3}\orcidID{1111-2222-3333-4444} \and
% Qibin Zhao\inst{3}\orcidID{2222--3333-4444-5555}}
%
\authorrunning{Z. Tao et al.}
\institute{Tokyo University of Agriculture and Technology, Tokyo, Japan \and
RIKEN AIP, Tokyo, Japan \\
\email{\{zerui.tao, qibin.zhao\}@riken.jp, tanakat@cc.tuat.ac.jp}}
\maketitle              % typeset the header of the contribution
\begin{abstract}
  Tensor decompositions play a crucial role in numerous applications related to multi-way data analysis.
  By employing a Bayesian framework with sparsity-inducing priors, Bayesian Tensor Ring (BTR) factorization offers probabilistic estimates and an effective approach for automatically adapting the tensor ring rank during the learning process.
  However, previous BTR \cite{long2021bayesian} method employs an Automatic Relevance Determination (ARD) prior, which can lead to sub-optimal solutions.
  Besides, it solely focuses on continuous data, whereas many applications involve discrete data.
  More importantly, it relies on the Coordinate-Ascent Variational Inference (CAVI) algorithm, which is inadequate for handling large tensors with extensive observations.
  These limitations greatly limit its application scales and scopes, making it suitable only for small-scale problems, such as image/video completion.
  To address these issues, we propose a novel BTR model that incorporates a nonparametric Multiplicative Gamma Process (MGP) prior, known for its superior accuracy in identifying latent structures.
  To handle discrete data, we introduce the \pg augmentation for closed-form updates.
  Furthermore, we develop an efficient Gibbs sampler for consistent posterior simulation, which reduces the computational complexity of previous VI algorithm by two orders, and an online EM algorithm that is scalable to extremely large tensors.
  To showcase the advantages of our model, we conduct extensive experiments on both simulation data and real-world applications.
\keywords{Tensor decomposition  \and Tensor completion \and Bayesian methods \and Gibbs sampler.}
\end{abstract}
\section{Introduction}

Multi-way data are ubiquitous in many real-world applications, such as recommender systems, knowledge graphs, images/videos and so on. Since multi-way arrays are typically high-dimensional and sparse, tensor decomposition (TD) serves as a powerful tool for analyzing such data.
Many elegant approaches for TD have been established, such as CANDECOM/PARAFAC (CP) \cite{hitchcock1927expression}, Tucker \cite{tucker1966some}, Tensor Train (TT) \cite{oseledets2011tensor}, Tensor Ring (TR) \cite{zhao2016tensor} and so on.
Among these TD models, TR decomposition and its descendants \cite{wang2017efficient,yuan2019tensor,long2021bayesian,tao2021bayesian} have achieved great successes, due to its highly compact form and expressive power. In particular, Bayesian TR (BTR) \cite{long2021bayesian,tao2021bayesian} has many advantages.
Firstly, by adopting fully probabilistic treatment, uncertainty estimates can be obtained, which is desirable for real-world applications.
Secondly, with sparsity-inducing priors, it establishes a principled way of automatic rank adaption during the learning process and greatly avoid over-fitting.
However, previous BTR are developed using Coordinate-Ascent Variational Inference (CAVI) \cite{long2021bayesian} or Expectation-Maximization (EM) \cite{tao2021bayesian} algorithms, which can not handle large tensors with massive observations. Besides, both \cite{long2021bayesian} and \cite{tao2021bayesian} did not consider discrete data, which are important in real-world applications.

To address the above issues, we proposed a novel BTR model which can deal with continuous or binary data and is scalable to large datasets. In particular, we firstly propose an weighted version of TR, which incorporates the nonparametric Multiplicative Gamma Process (MGP) prior \cite{bhattacharya2011sparse} for automatic rank adaption. The Gaussian assumption is then extended with \pg augmentation \cite{polson2013bayesian} to deal with binary data. To simulate the posterior, we develop an efficient Gibbs sampler, which has smaller computational complexity than the CAVI algorithm in \cite{long2021bayesian}. Moreover, an online version of EM algorithm is established to handle large tensors with massive observations. To showcase the advantages of our model, we conduct extensive experiments on both simulation data and real-world applications. The contributions are summarized as follows: (1) We propose a novel weighted TR decomposition with MGP prior. (2) The model is extended with \pg augmentation to deal with binary data. (3) An efficient Gibbs sampler and a scalable online EM algorithm is developed to deal with large datasets.

\section{Related Work}

To deal with multi-way data, many elegant tensor decompositions (TD) have been proposed, including CP \cite{hitchcock1927expression}, Tucker \cite{tucker1966some}, tensor train \cite{oseledets2011tensor}, tensor ring \cite{zhao2016tensor} and their descendants \cite{kolda2009tensor,cichocki2016tensor}. Traditionally, TD factors are learnt through alternating least square (ALS) algorithms \cite{kolda2009tensor,zhao2016tensor,wang2017efficient}. However, to handle large tensors with massive data, (stochastic) gradient-based optimization can also be applied \cite{acar2011scalable,oh2018scalable,yuan2017completion}. Despite their success, Bayesian TDs have many advantages over traditional ones, such as uncertainty estimates, adaptive rank selection, a principled way of modeling diverse data types. For example, \cite{zhao2015bayesian} proposed Bayesian CP to automatically select tensor ranks. \cite{rai2014scalable,rai2015scalable,cheng2018scaling} further derived online EM/VI algorithm for Bayesian CP to deal with both continuous and discrete data. Besides, it is also possible to derive Bayesian version of Tucker decomposition \cite{schein2016bayesian,zhao2015bayesiantucker}. However, CP and Tucker have their limitations. CP format lacks flexibility to deal with complex real-world data, while Tucker suffers from the curse of dimensionality for high-order tensors. Tensor ring (TR) decomposition has shown predominant performances in many application, such image processing \cite{wang2017efficient,yuan2019tensor}, model compression \cite{wang2018wide}, generative models \cite{kuznetsov2019prior} and so on. To equip TR format with Bayesian framework, \cite{long2021bayesian,tao2020bayesian} developed a Bayesian TR with Automatic Relevance Determination (ARD) prior for sparsity and derived a Coordinate-Ascent Variational Inference (CAVI) algorithm to learn posteriors. \cite{tao2021bayesian} proposed a Bayesian TR model for factor analysis, which employs Multiplicative Gamma Process (MGP) \cite{bhattacharya2011sparse} prior and EM algorithm. Other related works include Bayesian CP with generalized hyperbolic prior \cite{cheng2022towards}, Bayesian TT with Gaussian-product-Gamma prior \cite{xu2021probabilistic}. Most of these works are not scalable and cannot handle discrete data.

\section{Backgrounds}
\label{sec:background}

\subsection{Notations}

In machine learning community, the term ``tensor'' usually refers to multi-way arrays, which extends vectors and matrices. For consistency with previous literature, the style of notations follows the convention in \cite{kolda2009tensor}.
We use lowercase letters, bold lowercase letters, bold capital letters and calligraphic bold capital letters to represent scalars, vectors, matrices and tensors, \eg, $x, \ve{x}, \mat{x}$ and $\ten{x}$. For an order-$D$ tensor $\ten{x} \in \mathbb{R}^{I_1 \times \cdots \times I_D}$, we denote  its $(i_1, \dots, i_{D})$-th entry as $x_{\ind}$. Moreover, $\mathcal{N}(\cdot, \cdot)$ denotes Normal distribution, $Ga(\cdot, \cdot)$ denotes Gamma distributions and $\kl{\cdot}{\cdot}$ denotes the Kullback-Leibler (KL) divergence.

\subsection{Tensor Ring Decomposition}

In this subsection, we introduce the tensor ring (TR) decomposition \cite{zhao2016tensor}. Given an order-$D$ tensor $\ten{X} \in \bR^{I_1 \times \cdots \times I_D}$, TR format factorizes it into $D$ core tensors,
\begin{equation}\label{eq:origin_trd}
  x_{\mbi} = \tr\left( \mat{G}^{(1), i_1} \mat{G}^{(2), i_2} \cdots \mat{G}^{(D), i_D} \right),
\end{equation}
where $\mat{G}^{(d), i_d} \in \bR^{R_d \times R_d}, \forall i_d = 1, \dots, I_d$ are $i_{d}$-th slices of the core tensor $\ten{G}^{(d)} \in \bR^{I_d \times R_{d} \times R_{d}}$. And the sequence $\{R_d\}_{d=1}^{D}$ is the rank for a TR format. Typically, we assume $R=R_1=\cdots=R_{d}$ for convenience. We can denote a TR format as
\( \ten{X} = \mathit{TR}(\ten{G}^{(1)}, \dots, \ten{G}^{(D)}). \)
For ease of the derivation, we introduce the subchain of a TR format by contracting a subsequence of core tensors, namely,
\( \ten{G}^{\neq d} \in \bR^{I_{d+1} \times \cdots \times I_D \times I_1 \times I_{d-1} \times R \times R}, \)
where each slice becomes,
\begin{equation}\label{eq:tr-subchain}
 \mat{G}^{\neq d, i_{d+1}\cdots i_{d-1}} = \mat{G}^{(d+1), i_{d+1}} \cdots \mat{G}^{(d-1), i_{d-1}} \in \mathbb{R}^{R \times R}.
\end{equation}
Then the TR format can be simplified as \(x_{\mbi} = \tr\left( \mat{G}^{(d), i_d} \mat{G}^{\neq d, i_{d+1}\cdots i_{d-1}} \right).\)

\section{Proposed Model}

\subsection{Weighted Tensor Ring Decomposition}

To enable the rank inference, we employ a weight variable $\mat{\Lambda}$ that indicates the importance of each factor and rewrite \cref{eq:origin_trd} as,
\begin{equation}\label{eq:tr_model}
  x_{\mbi} = \tr\left( \mat{G}^{(1), i_1} \mat{\Lambda}^{(1)} \mat{G}^{(2), i_2} \mat{\Lambda}^{(2)} \cdots \mat{G}^{(D), i_D} \mat{\Lambda}^{(D)} \right),
\end{equation}
where $\mat{\Lambda}^{(d)} = \diag(\lambda^{(d)}_1, \dots, \lambda^{(d)}_R)$. Hence, if $\lambda^{(d)}_r$ becomes small, we can prune the corresponding redundant factor. Absorbing the weights into the core tensors, we can get an equivalent TR format
\( \ten{X} = \mathit{TR}(\ten{\tilde{G}}^{(1)}, \dots, \ten{\tilde{G}}^{(D)}), \)
where $\mat{\tilde{G}}^{(d), i_d} = \mat{G}^{(d), i_d} \mat{\Lambda}^{(d)}, \quad \forall i_d =1 , \dots, I_d.$
In real-world applications, the observations are typically corrupted by noises. In this paper, we consider both continuous and binary data, following Gaussian and Bernoulli distributions as follows.
(1) Continuous data: Given a set of observations $\Omega = \{\ind_i, \dots, \ind_{N}\}$, we assume the observations follow Gaussian distribution,
  \begin{equation}\label{eq:data-cont}
    p(\ten{Y}_{\Omega} \mid \ten{X}) = \prod_{\mbi \in \Omega} \mathcal{N}(y_{\mbi} \mid x_{\mbi}, \tau^{-1}),
  \end{equation}
where $\tau$ is the noise precision and can be inferred under Bayesian framework.
(2) Binary data: We employ the Bernoulli model,
  \begin{equation}\label{eq:data-binary}
    p(\ten{Y}_{\Omega} \mid \ten{X}) = \prod_{\mbi \in \Omega} \left(\frac{1}{1 + \exp(- x_\mbi)}\right)^{y_\mbi} \left( \frac{\exp(- x_\mbi)}{1 + \exp(- x_\mbi)} \right)^{1 - y_\mbi}.
  \end{equation}

\subsection{Probabilistic Model with Sparsity Inducing Prior}

\begin{figure}[t]
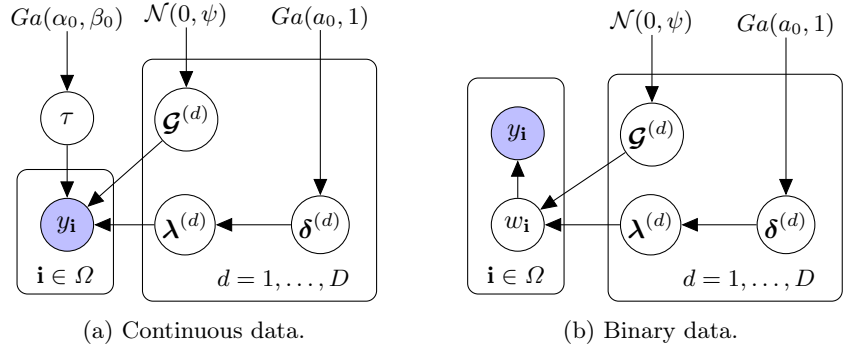

  \centering
  \hspace{-1em}
  \begin{subfigure}[t]{0.4\textwidth}
  % \centering
  \tikz{ %
    \node[obs] (y_obs) {$y_\mbi$};
    \node[latent, above=of y_obs, yshift=-0.3cm] (noise) {$\tau$} ; %
    \node[const, above=of noise, xshift=0.cm, yshift=-0.2cm] (noise_prior) {$Ga(\alpha_{0}, \beta_{0})$};
    \node[latent, right=of y_obs, yshift=1.4cm, xshift=-0.2cm] (core_q) {$\ten{G}^{(d)}$} ;
    \node[const, above=of core_q, xshift=0.cm, yshift=-0.2cm] (q_prior) {$\mathcal{N}(0, \psi)$} ;
    \node[latent, right=of y_obs, yshift=0.cm, xshift=-0.2cm] (core_lambda) {$\ve{\lambda}^{(d)}$} ;
    \node[latent, right=of core_lambda] (delta) {$\ve{\delta}^{(d)}$} ;
    \node[const, above=of delta, xshift=0.cm, yshift=1.2cm] (delta_prior) {$Ga(a_{0}, 1)$};

    \plate[inner sep=0.25cm, xshift=0.cm, yshift=0.12cm] {plate1} {(y_obs)} {$\mbi \in \Omega$} ; %
    \plate[inner sep=0.25cm, xshift=0.1cm, yshift=0.12cm] {plate2} {(core_q) (core_lambda) (delta)} {$d =1, \dots, D$} ; %
    % \plate[inner sep=0.05cm, xshift=0.cm, yshift=0.08cm] {plate3} {(delta)} {$r$} ; %
    % \plate[inner sep=0.05cm, xshift=0.cm, yshift=0.08cm] {plate4} {(psi)} {$r$} ; %

    \edge {noise} {y_obs} ;
    \edge {noise_prior} {noise};
    \edge {core_q} {y_obs} ;
    % \edge {psi_prior} {psi};
    \edge {q_prior} {core_q};
    \edge {core_lambda} {y_obs} ;
    \edge {delta} {core_lambda} ;
    \edge {delta_prior} {delta} ;
  }
  \caption{Continuous data.}
  \end{subfigure}
  % \hfill
  \hspace{3em}
  \begin{subfigure}[t]{0.4\textwidth}
  % \centering
  \tikz{ %
    \node[latent] (x_obs) {$w_\mbi$};
    \node[obs, above= of x_obs, yshift=-0.5cm] (y_obs) {$y_\mbi$};
    \node[latent, right=of x_obs, yshift=1.2cm] (core_q) {$\ten{G}^{(d)}$} ;
    \node[const, above=of core_q, xshift=0.cm, yshift=-0.1cm] (q_prior) {$\mathcal{N}(0, \psi)$} ;
    \node[latent, right=of x_obs, yshift=0.cm] (core_lambda) {$\ve{\lambda}^{(d)}$} ;
    \node[latent, right=of core_lambda] (delta) {$\ve{\delta}^{(d)}$} ;
    \node[const, above=of delta, xshift=0.cm, yshift=1.1cm] (delta_prior) {$Ga(a_0, 1)$};

    \plate[inner sep=0.25cm, xshift=0.0cm, yshift=0.12cm] {plate1} {(y_obs) (x_obs)} {$\mbi \in \Omega$} ; %
    \plate[inner sep=0.25cm, xshift=0.1cm, yshift=0.12cm] {plate2} {(core_q) (core_lambda) (delta)} {$d=1, \dots, D$} ; %

    \edge {x_obs} {y_obs}
    \edge {q_prior} {core_q};
    \edge {core_q, core_lambda} {x_obs} ;
    \edge {delta} {core_lambda} ;
    \edge {delta_prior} {delta} ;
  }
  \caption{Binary data.}
  \end{subfigure}
  \caption{Graphical models of Bayesian weighted tensor ring decomposition.}
  \label{fig:graph}
\end{figure}

To induce low-rank sparsity, we add a Multiplicative Gamma Process (MGP) prior \cite{bhattacharya2011sparse} on the weight parameters.
For $d = 1, \dots, D$ and $r = 1, \dots, R$, let
\[ \lambda^{(d)}_r \sim \mathcal{N}(0, (\phi_r^{(d)})^{-1}), \quad \delta^{(d)}_l \sim Ga(a_0, 1), \]
where $\phi^{(d)}_r = \prod_{l=1}^r \delta^{(d)}_l$ and $a_0 > 1$ to ensure sparsity.
Then core tensors are assigned Gaussian distribution,
\[ g^{(d), i_d}_{r, r'} \sim \mathcal{N}(0, (\psi^{(d), i_d}_{r, r'})^{-1}), \forall r, r' = 1, \dots, R, \]
where $g^{(d), i_d}_{r, r'}$ are $(r, r')$-th elements of core tensor $\mat{G}^{(d), i_{d}}$.
The precision can be either constant, \eg, $\psi^{(d), i_d}_{r, r'} = 1$, or follow Gamma prior, \( \psi^{(d), i_d}_{r, r'} \sim Ga(c_0, d_0). \)
Finally, for continuous data, the noise precision follows a Gamma distribution,
\( \tau \sim Ga(\alpha_0, \beta_0). \)
For binary data, we use an auxiliary variable $\ten{W}$ for data augmentation, for which the details are presented in \cref{sec:binary-tensor}.
The whole parameter set is denoted as $\mat{\Theta} = \{\tau, \ten{G}, \mat{\Lambda}, \ve{\delta}, \ten{w} \}$.
Integrating all the prior distributions, we can get the joint prior distribution.
For example, for continuous data,
\begin{equation}\label{eq:joint}
  \begin{multlined}[0.8\textwidth]
  p(\mat{\Theta}) = Ga(\tau \mid \alpha_0, \beta_0) \cdot \prod_{d=1}^D \prod_{i_d=1}^{I_{d}} \prod_{r, r'=1}^{R} \mathcal{N}(g^{(d), i_{d}}_{r, r'} \mid 0, (\psi^{(d), i_{d}}_{r, r'})^{-1}) \\
  \cdot \prod_{d=1}^D \prod_{r}^R \mathcal{N}(\lambda^{(d)}_r \mid 0, (\phi^{(d)}_r)^{-1}) \cdot Ga(\delta^{(d)}_r \mid a_0, 1).
  \end{multlined}
\end{equation}
Then we can construct the whole generative model as,
\[ p(\ten{Y}, \mat{\Theta}) = p(\ten{Y} \mid \ten{x}) \cdot p(\tau, \ten{G}, \mat{\Lambda}, \delta), \]
where $p(\ten{Y} \mid \ten{x})$ can be either Gaussian distribution \cref{eq:data-cont} or Bernoulli distribution \cref{eq:data-binary}.
\cref{fig:graph} shows the graphical illustration of the probabilistic model.
% The graphical model of the proposed model is shown in \cref{fig:graph}.

\section{Gibbs Sampler}
\label{sec:gibbs}

\subsection{Sampling Rules}

The basic idea of Gibbs sampler is to sequentially sample each parameter $\theta_{j}$ from conditional distribution $p(\theta_j \mid \theta_{1}, \dots, \theta_{-j})$, where $\theta_{-j}$ means parameters exclude $\theta_{j}$. It can be proved that the stationary distribution of this Markov chain is the true posterior.
Due to the conjugate priors, all the conditional distributions can be derived analytically.
More details can be find in tutorials such as \cite{andrieu2003introduction}.

\paragraph{Sample $\ve{\delta}$.}
To update the auxiliary variable $\ve{\delta}$, we sample from the conditional distribution
\begin{equation}\label{eq:gibbs_delta}
  \delta^{(d)}_r \mid - \sim Ga\left( a_0 + \frac{1}{2}(R - r + 1), 1 + \frac{1}{2}\sum_{h=r}^R \lambda^{(d), r}_{h} \prod_{l=1, l\neq r}^h \delta^{(d)}_l \right) .
\end{equation}

\paragraph{Sample $\ve{\lambda}$.}
To update $\lambda$, we firstly rewrite the TR model \cref{eq:tr_model} as
\begin{equation*}
    x_\mbi = \tr \left(\mat{\Lambda}^{(d)} \cdot \mat{\tilde{G}}^{\neq d, \mbi_{-d}} \cdot \mat{G}^{(d), i_d} \right) = a^r_\mbi \lambda^{(d)}_r + b^r_\mbi,
\end{equation*}
where
\(a^r_\mbi = \left(\mat{\tilde{G}}^{\neq d, \mbi_{-d}} \cdot \mat{G}^{(d), i_d} \right)_{rr}\) and \(b^r_\mbi = \sum_{r' \neq r} \left(\mat{\tilde{G}}^{\neq d, \mbi_{-d}} \cdot \mat{G}^{(d), i_d} \right)_{r'r'} \lambda^{(d)}_{r'}\), with \(\mbi_{-d}\) denoting \([i_{d+1}, \dots, i_{D}, i_{1}, \dots, i_{d=1} ] \).
Then we can sample from the conditional distribution,
\begin{equation}\label{eq:gibbs_lambda}
  \lambda^{(d)}_r \mid - \sim \mathcal{N}(\mu^{(d)}_r, (\sigma^{(d)}_r)^{-1}),
\end{equation}
where
\(\mu^{(d)}_r = (\sigma^{(d)}_r)^{-1} \tau \sum_\mbi a^r_\mbi (y_\mbi - b^r_\mbi) \), and  \(\sigma^{(d)}_r = \phi^{(d)}_r + \tau \sum_\mbi (a^r_\mbi)^2\).

\paragraph{Sample core tensors $\ten{G}$.}
The derivation is similar with that in updating $\ve{\lambda}$. Particularly, the TR model \cref{eq:tr_model} is rewritten as
\begin{equation*}
    x_\mbi = \tr \left( \mat{G}^{(d), i_d} \cdot \mat{\Lambda}^{(d)} \cdot \mat{\tilde{G}}^{\neq d, \mbi_{-d}} \right) = c^{(k), i_d}_{r, r'} g^{(d), i_d}_{r, r'} + d^{(d), i_d}_{r, r'},
\end{equation*}
where
\(c^{(k), i_d}_{r, r'} = \left(\mat{\Lambda}^{(d)} \cdot \mat{\tilde{G}}^{\neq d, \mbi_{-d}} \right)_{rr'}\) and \(d^{(d), i_d}_{r, r'} = \sum_{(i, j)\neq (r, r')}^R \left( \mat{\Lambda}^{(d)} \mat{\tilde{G}}^{\neq d, \mbi_{-d}} \right)_{ij} g^{(d), i_d}_{j,i}\).
Then the conditional distribution becomes,
\begin{equation}\label{eq:gibbs_core}
  \mat{G}^{(d)}_{r, r'} \mid - \sim \mathcal{N}(\ve{\mu}^{(d)}_{r, r'}, \mat{\Sigma}^{(d)}_{r, r'}).
\end{equation}
The variance is computed by
\begin{equation}
\label{eq:gibbs-sigma}
  \mat{\Sigma}^{(d)}_{r, r'} = (\diag(\psi^{(d)}_{r, r'}) + \mat{T}^{(d)}_{r, r'})^{-1},
\end{equation}
where \(\mat{T}^{(d)}_{r, r'} = \diag( t^{(d), i_1}_{r, r'}, \dots, t^{(d), I_d}_{r, r'})\) and \(t^{(d), i_d}_{r, r'} = \tau \sum_{\mbi, i = i_d} (c^{(d), i}_{r, r'})^2\).
Additionally, the mean parameter is computed as
\begin{equation}\label{eq:gibbs-mu}
  \ve{\mu}^{(d)}_{r, r'} = \mat{\Sigma}^{(d)}_{r, r'} \mat{T}^{(d)}_{r, r'} \ve{\alpha}^{(d)}_{r, r'},
\end{equation}
where \( \ve{\alpha}^{(d)}_{r, r'} = [ \alpha^{(d), i_d}_{r, r'}, \dots,  \alpha^{(d), I_d}_{r, r'}] \), \(\alpha^{(d), i_d}_{r, r'} = (t^{(d), i_d}_{r, r'})^{-1} \tau \sum_{\mbi, i = i_d} c^{(k), i}_{r, r'} (y_\mbi - d^{(k), i}_{r, r'})\).

\paragraph{Sample $\tau$.}
To update the noise precision $\tau$, we can simply sample from the Gamma distribution
\begin{equation}\label{eq:gibbs_noise}
  \tau \mid - \sim Ga\left( \alpha_{0} + \frac{\lvert \Omega \lvert}{2}, \beta_{0} + \frac{1}{2} \sum_{\mbi \in \Omega}( y_\mbi - x_\mbi )^2 \right),
\end{equation}
where $\lvert \Omega \lvert$ is the number of observed entries and $\alpha_{0}, \beta_{0}$ are hyperparameters.

\subsection{Binary Tensor}\label{sec:binary-tensor}
To deal with binary tensor, we employ the \pg (PG) augmentation \cite{polson2013bayesian} by introducing the auxiliary variable $\ten{w}$, which follows the PG distribution,
\begin{equation}
\label{eq:pg-binary}
\omega_\mbi \sim PG(1, x_\mbi).
\end{equation}
Due to properties of PG distribution, we can sample $\ve{\lambda}$ from
\begin{equation}\label{eq:sample-lambda-binary}
  \lambda^{(d)}_r \mid - \sim \mathcal{N}(\mu^{(d)}_r, (\sigma^{(d)}_r)^{-1}), \quad \forall r = 1, \dots, R,
\end{equation}
where
\(\mu^{(d)}_r = (\sigma^{(d)}_r)^{-1} \sum_\mbi a^r_\mbi (y_\mbi - 0.5 - w_\mbi b^r_\mbi) \) and \( \sigma^{(d)}_r = \phi^{(d)}_r + \sum_\mbi (a^r_\mbi)^2 w_\mbi\).
The sampling rule of core tensors $\ten{g}$ becomes,
\begin{equation}\label{eq:sample-q-binary}
  \mat{G}^{(d)}_{r, r'} \mid - \sim \mathcal{N}(\ve{\mu}^{(d)}_{r, r'}, \mat{\Sigma}^{(d)}_{r, r'}),
\end{equation}
where \( \mat{\Sigma}^{(d)}_{r, r'} \) and \( \ve{\mu}^{(d)}_{r, r'} \) are computed by \cref{eq:gibbs-sigma} and \cref{eq:gibbs-mu} respectively, with \(t^{(d), i_d}_{r, r'} = \sum_{\mbi, i = i_d} (c^{(d), i}_{r, r'})^2 w_\mbi\) and \(\alpha^{(d), i_d}_{r, r'} = (t^{(d), i_d}_{r, r'})^{-1} \sum_{\mbi, i = i_d} c^{(k), i}_{r, r'} (y_\mbi - 0.5 - w_\mbi d^{(k), i}_{r, r'})\).
For binary data, there is no need to sample $\tau$. The overall procedure is presented in \cref{algo:gibbs}.

\subsection{Tensor Ring Rank Adaption}\label{sec:rank-adaption}

In TD problems, one essential issue is choosing the appropriate rank.
In Bayesian framework, the presence of sparsity-inducing priors allows us to adaptively search ranks during training.
Previous BTR-VI model with ARD prior \cite{long2021bayesian} proposed to start with a large TR rank $R_{0}$, and then prune redundant factors during training, which we refer to \emph{truncation} strategy.
Such strategy has two main limitations.
Firstly, if we choose a large $R_{0}$, the computational complexity becomes extremely large at beginning.
Secondly, if the predefined $R_{0}$ is smaller than the true rank, we can never learn it.
Due to the nonparametric property of the MGP prior, the proposed model can potentially learn infinite number of latent factors.
To achieve this, we adopt the \emph{adaption} strategy, which is similar with \cite{bhattacharya2011sparse,rai2014scalable}.
In particular, the model starts with a relatively smaller $R_{0}$.
Then, after each iteration, we prune factors with weights smaller than a threshold, e.g., $\lambda^{(d)}_{r} < \epsilon$.
If all of the weights are larger than the threshold, a new factor is added with a probability $p(t) = \exp(\kappa_0 + \kappa_{1} t)$, where $t$ is the current iteration number.
The \emph{adaption} strategy has much lower computational burden at the beginning stage than the \emph{truncation} strategy, and can learn higher ranks than the initial rank.

\begin{algorithm}[t]
\caption{Gibbs sampler}
\label{algo:gibbs}
\begin{algorithmic}[1]
\STATE Input observation $\ten{y}_{\Omega}$ and hyperparameters $a_{0}, \alpha_0, \beta_{0}$.

\FOR{$t \leq max\_step$ }

\FOR{$d = 1, \dots, D$, $r = 1, \dots, R$}
\STATE Sample $\delta^{(d)}_r$ from \cref{eq:gibbs_delta}. \COMMENT{Sample $\delta$}
\ENDFOR

\FOR{$d = 1, \dots, D$, $r = 1, \dots, R$}
\STATE Sample $\lambda^{(d)}_r$ from \cref{eq:gibbs_lambda} or \cref{eq:sample-lambda-binary}. \COMMENT{Sample $\lambda$}
\ENDFOR

\FOR{$d = 1, \dots, D$, $r = 1, \dots, R$, $r' = 1, \dots, R$}
\STATE Sample $\mat{G}^{(d)}_{r, r'}$ from \cref{eq:gibbs_core} or \cref{eq:sample-q-binary}. \COMMENT{Sample $\ten{G}$}
\ENDFOR

\IF{$\ten{y}$ is continuous}
\STATE Sample $\tau$ from \cref{eq:gibbs_noise}. \COMMENT{Sample $\tau$}
\ELSIF{$\ten{Y}$ is binary}
\FOR{$\mbi$ in $\Omega$}
\STATE Sample $w_{\mbi}$ from \cref{eq:pg-binary}. \COMMENT{Sample $\ten{w}$}
\ENDFOR
\ENDIF

\STATE Rank adaption as described in \cref{sec:rank-adaption}. \COMMENT{Rank adaption}

\ENDFOR
\end{algorithmic}
\end{algorithm}

\subsection{Complexity Analysis}

We denote the tensor rank as $R$, the tensor order as $D$ and the number of observations as $M$.
Computing \cref{eq:gibbs_delta} has complexity $\mathcal{O}(R^2)$, hence the complexity of sampling $\delta$ is $\mathcal{O}(DR^3)$. Computing \cref{eq:gibbs_lambda} has complexity $\mathcal{O}(MR^2)$, hence the complexity of sampling $\delta$ is $\mathcal{O}(DMR^3)$. Computing \cref{eq:gibbs_core} has complexity $\mathcal{O}(MR^2)$, hence the complexity of sampling $\delta$ is $\mathcal{O}(DMR^4)$. The complexity of updating $\epsilon$ is $\mathcal{O}(M)$ and can be omitted. Hence, the overall complexity is $\mathcal{O}(DMR^4)$, which is smaller than the BTR-VI \cite{long2021bayesian} ($\mathcal{O}(DMR^{6})$) by two orders.

\section{Online EM Algorithm}

In real applications, tensors can be very large with massive observations.
For such cases, the Gibbs sampler that uses the whole dataset is prohibited.
To address the issue, we establish an online Variational Bayes EM (VBEM) algorithm that employs stochastic updates which is scalable to large tensors.

For illustration, we consider the binary case. The continuous case can be derived straightforwardly.
Recall that the full probabilistic is
\begin{equation*}
  p(\ten{Y}, \Theta) = p(\ten{Y} \mid \ten{W}) p(\ten{W} \mid \ten{G}, \ve{\Lambda}) p(\ve{\Lambda} \mid \delta) p(\delta) p(\ten{G}).
\end{equation*}
To enable the VBEM algorithm, we assign a variational distribution \(
  q(\ten{W}) q(\delta) \approx p(\ten{W}, \delta \mid \ten{Y}, \mat{\Lambda}, \ten{G}),
\)
with parametric forms as follows,
\begin{equation*}
    q(\ten{W}) = \prod_{\mbi} PG(1, x_\mbi), \quad q(\delta) = \prod_{d=1}^D \prod_{r=1}^R Ga(\alpha^{(d), r}_\delta, \beta^{(d), r}_\delta).
\end{equation*}
In the VBEM algorithm, we derive closed-form expectations of \(\{ \ten{w}, \delta \}\), and then update the rest parameters by gradient ascent of the free energy.

\paragraph{E-step.}

Using similar derivation with \cref{sec:gibbs}, we have the expectations,
\begin{equation}\label{eq:em-exp}
  \E_q [\delta^{(d)}_r] = \frac{a_\delta + \frac{1}{2}(R - r + 1)}{1 + \frac{1}{2}\sum_{h=r}^R \lambda^{(d), r}_{h} \prod_{l=1, l\neq r}^h \delta^{(d)}_l}, \quad
  \E_q [w_\mbi] = \frac{1}{2 x_\mbi} \mathrm{tanh}(x_\mbi / 2).
\end{equation}

\paragraph{M-step.}
Parameters $\{ \ten{G}, \ve{\Lambda} \}$ are then updated by maximizing the free energy,
\begin{equation*}
  \begin{aligned}
    \mathcal{L} =& \E_{q} [ \log p(\ten{Y}, \Theta) ] \\
    =& \sum_{\mbi \in \Omega} \E_q [\log p(y_\mbi \mid w_\mbi) + \log PG(w_\mbi \mid 1, x_\mbi)] + \sum_{d=1}^D \sum_{r, r'=1}^{R} \log \mathcal{N}(\mat{G}^{(d)}_{r, r'} \mid \ve{0}, \mat{I}) \\
    &+ \sum_{d=1}^D \sum_{r=1}^{R} \E_q [ \log \mathcal{N}(\lambda^{(d)}_r \mid 0, (\phi_r^{(d)})^{-1})] + \sum_{d=1}^{D} \sum_{r=1}^{R} \E_{q} [\log Ga(\delta^{(d)}_r \mid \alpha_{0}, \beta_{0}) ],
  \end{aligned}
\end{equation*}
where the expectations can be computed via \cref{eq:em-exp}.

Since the free energy is factorized over observations $\mbi$, stochastic optimization with mini-batch samples can be adopted.
The gradient can be computed through either analytical expressions or back propagation.
The complexity of computing the free energy is $\mathcal{O}(BDR^{4})$, where $B$ is the mini-batch size.
This complexity is the same order with the Gibbs sampler.
However, when applying the online EM algorithm, we can choose small mini-batch size $B$ to handle large-scale problems.

% \paragraph{Complexity Analysis.}

\begin{figure}[t]
  \captionsetup[subfigure]{font=scriptsize} % Set font size for subfigure captions
  \begin{minipage}{0.5\textwidth}
    \centering
    \begin{subfigure}[b]{\linewidth}
      \includegraphics[width=\linewidth]{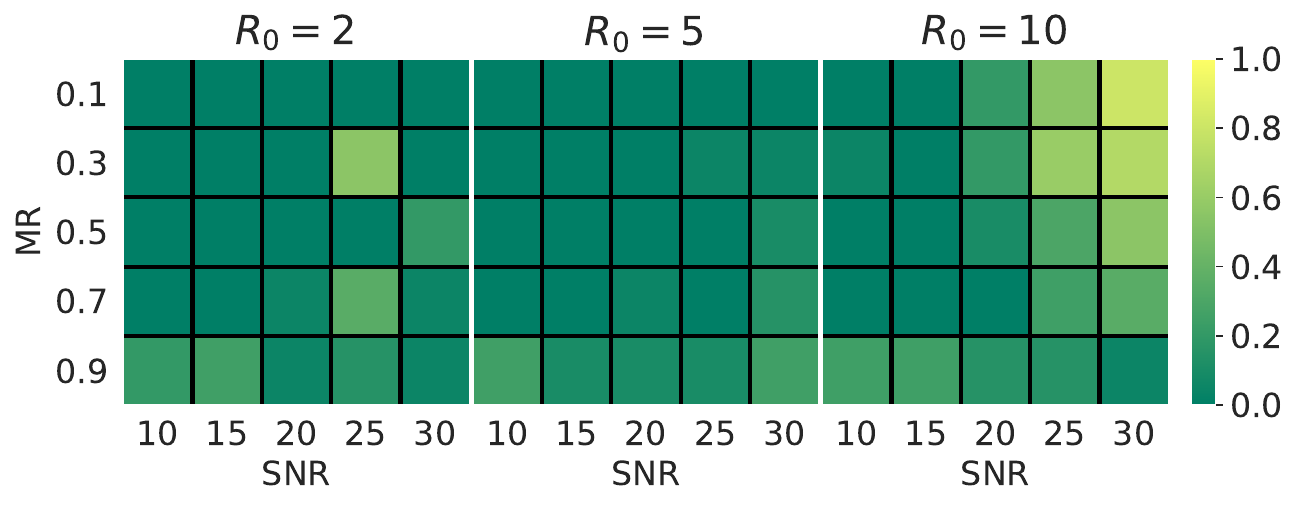}
      \caption{SBTR-Gibbs}\label{fig:simulation-rank-gibbs}
    \end{subfigure}

    \begin{subfigure}[b]{\linewidth}
      \includegraphics[width=\linewidth]{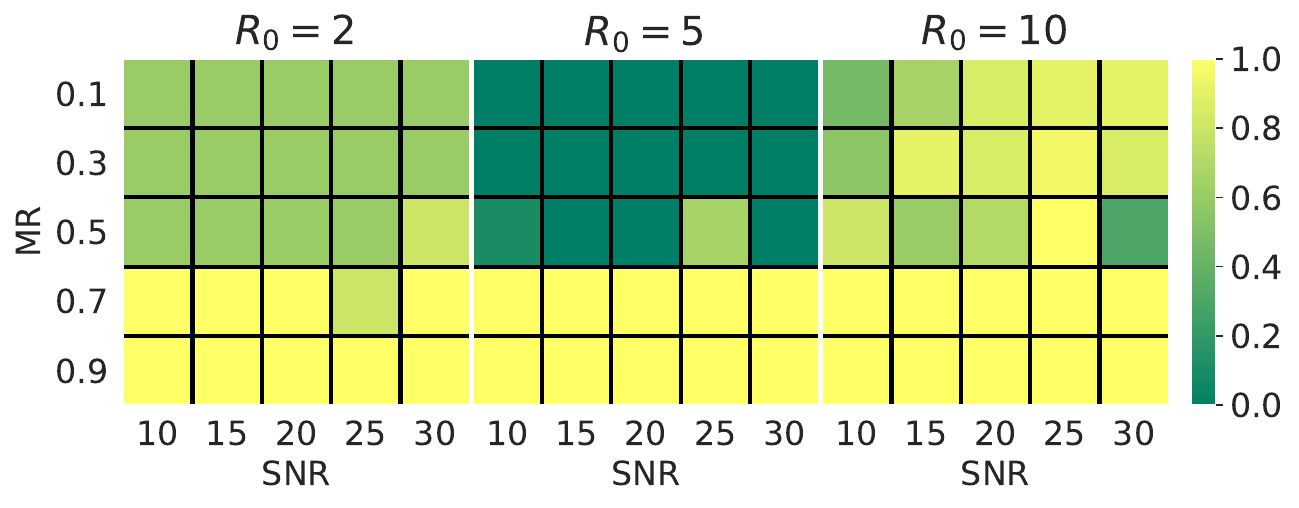}
      \caption{BTR-VI}\label{fig:simulation-rank-vi}
    \end{subfigure}
    \caption{Rank estimation results.}\label{fig:simulation-rank}
  \end{minipage}\hfill
  \begin{minipage}{0.5\textwidth}
    \centering
    \includegraphics[width=.9\linewidth]{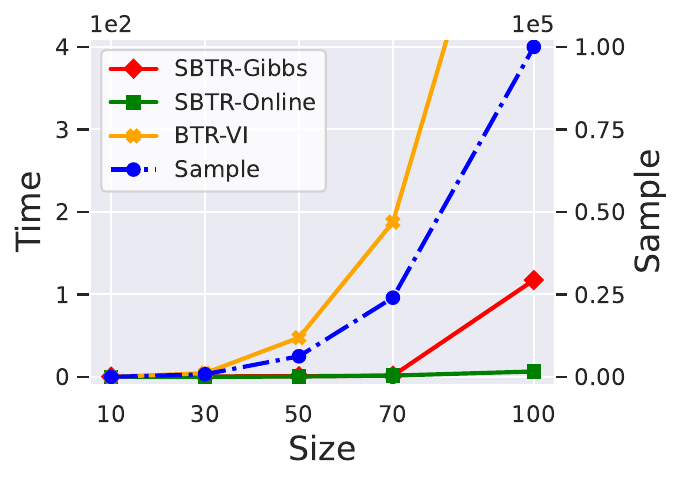}
    \caption{Scalability results.}\label{fig:simulation_time}
  \end{minipage}
\end{figure}

\section{Experiments\footnote{
The code is available at \url{https://github.com/taozerui/scalable_btr}
}}

In this section, we evaluate the proposed Scalable Bayesian Tensor Ring (SBTR) model on synthetic data and several real-world datasets. The Gibbs sampler and online EM algorithm are denoted as SBTR-Gibbs and SBTR-Online, respectively. All the experiments are conducted on a Linux workstation with Intel Xeon Silver 4316 CPU@2.30GHz, 512GB RAM and an NVIDIA RTX A6000 GPU (48GB memory). Our model is implemented based on PyTorch. If not specified, we run SBTR on GPU and other baselines on CPU.

\subsection{Simulation Study}

% First, we evaluate our model on tasks of rank estimation and scalability.

\paragraph{Rank Estimation.}

Firstly, we conduct simulation study on Tensor Ring (TR) rank estimation.
To show advantages of the MGP prior and Gibbs sampler, we compare with BTR-VI \cite{long2021bayesian}, which adopts ARD prior and variational inference.
We consider the true low rank tensor $\ten{x} \in \mathbb{R}^{10 \times 10 \times 10 \times 10}$ with TR rank 5.
We firstly generate TR core tensors from standard Gaussian distribution, then standardize $\ten{x}$ to have zero mean and unit variance.
Finally, i.i.d. Gaussian noises and uniform masks are added on the underlying signal $\ten{x}$.
The signal-to-noise ratio (SNR) is chosen from $\{10, 15, 20, 25, 30\}$ and the missing rates (MR) vary from $\{ 0.1, 0.3, 0.5, 0.7, 0.9 \}$.
For BTR-VI, we use their default settings and run 500 iterations.
For SBTR-Gibbs, we set hyperparameters $a_0 = 2.0, \alpha_0 = 1.0, \beta = 0.3$ and run 1500 burn-in steps to ensure convergence.
The rank estimation performance is evaluated using the relative absolute error $\mathrm{Err} = \sum_{d=1}^{4}\lvert \hat{R}_d - R_d \lvert / \sum_{d=1}^4 R_d$, where $\hat{R}_d$ is the estimated rank on each mode.

\cref{fig:simulation-rank} illustrates results under different initial rank $R_0$.
SBTR-Gibbs is consistently better than BTR-VI, especially when the missing ratio becomes large.

\paragraph{Scalability.}

Then, we showcase the scalability of our model.
In the experiment, an order-4 tensor of shape $I \times I \times \times I \times I$ is considered,
where the tensor size $I$ varies from $\{ 10, 30, 50, 70, 100 \}$ and the missing rate is set to $99.9\%$.
We compare the running time of BTR-VI, SBTR-Gibbs and SBTR-Online with TR rank 2.
All the models are tested on the CPU in this experiment.

The results are presented in \cref{fig:simulation_time}.
The solid lines are average time costs of one iteration (or, epoch for SBTR-Online) and the dashed line plots the sample sizes.
It shows that the BTR-VI algorithm does not scale well with the tensor size.
On the contrary, both SBTR-Gibbs and SBTR-Online show superior scalability.
The online EM algorithm is much faster than Gibbs sampler, since it can fully utilize parallel computation when choosing small batch sizes.

\subsection{Continuous Data Completion}\label{sec:cont-completion}

\paragraph{Datasets.}

We test our model on three continuous tensor for completion.
(1) U.S. Historical Climatology Network (USHCN)\footnote{\url{https://www.ncei.noaa.gov/products/land-based-station/us-historical-climatology-network}}, a climate dataset with shape $17 \times 125 \times 156$. We randomly select $10\%$ points for training and predict the rest entries.
(2) Indian Pines\footnote{\url{https://www.ehu.eus/ccwintco/index.php/Hyperspectral_Remote_Sensing_Scenes}}, a hyperspectral  image with shape $145 \times 145 \times 200$. We randomly set $99\%$ entries as missing and evaluate image completion performance.
(3) Alog \cite{zhe2015scalable}, a three-mode (\emph{user} \(\times\) \emph{action} \(\times\) \emph{resource}) tensor of shape $200 \times 100 \times 200$, which is extracted from an access log of a file management system. This data is partially observed with about $0.33\%$ nonzero entries. We use the same train/test split as in \cite{zhe2015scalable}.
All the datasets are evaluated using 5-fold cross-validation with different random seeds.

\begin{table}[t]
  \centering
  \small
  % \scriptsize
  \caption{Continuous tensor completion results. \red{Red} and \blue{Blue} entries indicate the best and second best, respectively.}\label{tab:cont-completion}
  \begin{tabular}{l|cc|cc|cc}
    \toprule
    & \multicolumn{2}{c|}{USHCN} &  \multicolumn{2}{c|}{Indian Pines} & \multicolumn{2}{c}{Alog}          \\
    \cmidrule(r){2-7}
    Model & RMSE$\downarrow$ & MAE$\downarrow$ & PSNR$\uparrow$ & SSIM$\uparrow$ & RMSE$\downarrow$ & MAE$\downarrow$ \\
    \midrule
    BayesCP & 0.590$\pm$0.006 & \blue{0.430$\pm$0.005} & 9.88$\pm$0.00 & 0.01$\pm$0.00 & 1.02$\pm$0.06 & 0.52$\pm$0.02 \\
    TTWOPT & 0.761$\pm$0.001 & 0.570$\pm$0.001 & 27.13$\pm$0.07 & 0.81$\pm$0.00 & 2.84$\pm$1.38 & 0.80$\pm$0.06 \\
    BayesTR-VI & 0.614$\pm$0.030 & 0.451$\pm$0.025 & 27.13$\pm$0.77 & 0.79$\pm$0.02 & 1.00$\pm$0.04 & 0.52$\pm$0.03 \\
    % \midrule
    SBTR-Gibbs & \blue{0.587$\pm$0.007} & \blue{0.430$\pm$0.003} & \red{29.71$\pm$0.12} & \red{0.84$\pm$0.00} & \blue{0.98$\pm$0.05} & \blue{0.51$\pm$0.02} \\
    SBTR-Online & \red{0.560$\pm$0.004} & \red{0.405$\pm$0.004} & \blue{28.07$\pm$0.04} & \blue{0.83$\pm$0.00} & \red{0.92$\pm$0.03} & \red{0.47$\pm$0.01} \\
    \bottomrule
  \end{tabular}
\end{table}

\paragraph{Competing methods.}

We compare with the following baseline models. (1) Bayesian CP (BayesCP) \cite{zhao2015bayesian}, a CP decomposition adopting ARD prior and variational inference. (2) Tensor Train Weighted OPTimization (TTWOPT) \cite{yuan2017completion}, which uses (stochastic) gradient descent to learn TT decomposition. (3) Bayesian TR with variational inference (BTR-VI) \cite{long2021bayesian} and ARD prior.

\paragraph{Experimental details and results.}

The most important hyperparameter is the tensor rank.
For BayesCP and BTR-VI, we set a large initial rank and prune redundant factors during training.
For SBTR-Gibbs, we use the rank adaption strategy described in \cref{sec:rank-adaption}.
The ranks of TTWOPT and SBTR-Online are chosen from \(\{3, 5, 10\}\) via validation.
Other settins of baselines are default in their codebases.
The hyperparameters of SBTR-Gibbs are $a_0 = 2.0, \alpha_0 = 1.0, \beta = 0.3$.
Moreover, we set batch size 512 and use Adam optimizer with learning rate 0.01 for SBTR-Online.

The completion results are shown in \cref{tab:cont-completion}.
For USHCN and Alog, we present the Root-Mean-Square Error (RMSE) and Mean Absolute Error (MAE), while for Indian, Peak Signal-to-Noise Ratio (PSNR) and Structural SIMilarity index (SSIM) are reported, since they are widely used to evaluate visual tasks.
The proposed SBTR model achieve the best or second best results in all tasks.

\subsection{Binary Data Completion}

\paragraph{Datasets.}

We evaluate our model on three binary datasets.
(1) Kinship \cite{nickel2011three}, which is a multirelational data of shape $104 \times 104 \times 26$, representing 26 relations of 104 entities.
(2) Enron \cite{zhe2015scalable}, a tensor of shape $203 \times 203 \times 200$ extracted from the Enron email dataset.
(3) DBLP \cite{zhe2015scalable}, a tensor of shape $10000 \times 200 \times 10000$ depicting bibliography relationships in the DBLP website.
Since these tensors are extremely sparse, we randomly sample the same number of zero and non-zeros entries to make the dataset balanced.
Then the results are evaluated using 5-fold cross validation with different random seeds.

\paragraph{Experimental details and results.}

\begin{table}[t]
  \centering
  % \small
  % \footnotesize
  \scriptsize
  \caption{Binary tensor completion results. \red{Red} and \blue{Blue} entries indicate the best and second best, respectively.}\label{tab:bin-completion}
  \begin{tabular}{l|cc|cc|cc}
    \toprule
    & \multicolumn{2}{c|}{Kinship} &  \multicolumn{2}{c|}{Enron} & \multicolumn{2}{c}{DBLP}          \\
    \cmidrule(r){2-7}
    Model & AUC$\uparrow$ & ACC$\uparrow$ & AUC$\uparrow$ & ACC$\uparrow$ & AUC$\uparrow$ & ACC$\uparrow$ \\
    \midrule
    BayesCP & \red{0.973$\pm$0.003} & \blue{0.925$\pm$0.012} & 0.500$\pm$0.000 & 0.500$\pm$0.000 & NA & NA \\
    TTWOPT & 0.943$\pm$0.008 & 0.774$\pm$0.156 & 0.819$\pm$0.049 & \blue{0.787$\pm$0.017} & 0.912$\pm$0.004 & 0.809$\pm$0.051 \\
    BayesTR-VI & \blue{0.968$\pm$0.010} & 0.896$\pm$0.024 & 0.500$\pm$0.000 & 0.500$\pm$0.000 & NA & NA \\
    % \midrule
    SBTR-Gibbs & 0.962$\pm$0.002 & 0.921$\pm$0.002 & \blue{0.834$\pm$0.018} & 0.758$\pm$0.019 & \blue{0.925$\pm$0.003} & \blue{0.869$\pm$0.002} \\
    SBTR-Online & \red{0.973$\pm$0.003} & \red{0.940$\pm$0.005} & \red{0.911$\pm$0.039} & \red{0.811$\pm$0.042} & \red{0.957$\pm$0.001} & \red{0.889$\pm$0.003} \\
    \bottomrule
  \end{tabular}
\end{table}

The settings are the same with those in \cref{sec:cont-completion}.
For baselines, we rescale their estimates to $[0, 1]$ as final predictions.
The prediction Area Under Curve (AUC) and ACCuracy (ACC) are presented in \cref{tab:bin-completion}.
In this experiment, SBTR-Online outperforms all other competitors.
For Enron dataset, ALS-based baselines, including BayesCP and BayesTR-VI are unable to get faithful predictions, while our SBTR-Gibbs gets much better results.
For DBLP dataset, BayesCP and BayesTR-VI exceed the maximum RAM of 512GB on our workstation.
Meanwhile, SBTR-Gibbs, which is also an ALS-based algorithm using the whole dataset, costs only around 1.2GB memory with Float32 precision on GPU.
All the results manifest the superior performance and scalability of the proposed SBTR.

\section{Conclusion}

This paper introduces a novel scalable Bayesian tensor ring factorization.
To achieve this, a weighted tensor ring factorization is proposed.
Then, a nonparametric multiplicative Gamma process is used to model the weights,
enabling automatic rank adaption during training.
A Gibbs sampler was proposed to obtained ALS-based updates, whose complexity is smaller than previous variational inference by two orders.
Moreover, an online EM algorithm was established to scale to large datasets.
We evaluate the proposed model on synthetic data, continuous tensor and binary tensor completion tasks.
The simulation results show the superior performance of our model in rank estimation and scalability.
Our model also outperforms all baselines in real-world tensor completion tasks.

\subsubsection{Acknowledgements}

This work was supported by the JSPS KAKENHI [Grant Number 20H04249, 23H03419].
Zerui Tao was supported by the RIKEN Junior Research Associate Program.

\bibliographystyle{splncs04}
\bibliography{ref.bib}

\begin{thebibliography}{10}
\providecommand{\url}[1]{\texttt{#1}}
\providecommand{\urlprefix}{URL }
\providecommand{\doi}[1]{https://doi.org/#1}

\bibitem{acar2011scalable}
Acar, E., Dunlavy, D.M., Kolda, T.G., M{\o}rup, M.: Scalable tensor
  factorizations for incomplete data. Chemometrics and Intelligent Laboratory
  Systems  \textbf{106}(1),  41--56 (2011)

\bibitem{andrieu2003introduction}
Andrieu, C., De~Freitas, N., Doucet, A., Jordan, M.I.: An introduction to mcmc
  for machine learning. Machine learning  \textbf{50},  5--43 (2003)

\bibitem{bhattacharya2011sparse}
Bhattacharya, A., Dunson, D.B.: Sparse bayesian infinite factor models.
  Biometrika  \textbf{98}(2),  291--306 (2011)

\bibitem{cheng2022towards}
Cheng, L., Chen, Z., Shi, Q., Wu, Y.C., Theodoridis, S.: Towards flexible
  sparsity-aware modeling: Automatic tensor rank learning using the generalized
  hyperbolic prior. IEEE Transactions on Signal Processing  \textbf{70},
  1834--1849 (2022)

\bibitem{cheng2018scaling}
Cheng, L., Wu, Y.C., Poor, H.V.: Scaling probabilistic tensor canonical
  polyadic decomposition to massive data. IEEE Transactions on Signal
  Processing  \textbf{66}(21),  5534--5548 (2018)

\bibitem{cichocki2016tensor}
Cichocki, A., Lee, N., Oseledets, I., Phan, A.H., Zhao, Q., Mandic, D.P.,
  et~al.: Tensor networks for dimensionality reduction and large-scale
  optimization: Part 1 low-rank tensor decompositions. Foundations and
  Trends{\textregistered} in Machine Learning  \textbf{9}(4-5),  249--429
  (2016)

\bibitem{hitchcock1927expression}
Hitchcock, F.L.: The expression of a tensor or a polyadic as a sum of products.
  Journal of Mathematics and Physics  \textbf{6}(1-4),  164--189 (1927)

\bibitem{kolda2009tensor}
Kolda, T.G., Bader, B.W.: Tensor decompositions and applications. SIAM review
  \textbf{51}(3),  455--500 (2009)

\bibitem{kuznetsov2019prior}
Kuznetsov, M., Polykovskiy, D., Vetrov, D.P., Zhebrak, A.: A prior of a googol
  gaussians: a tensor ring induced prior for generative models. Advances in
  Neural Information Processing Systems  \textbf{32} (2019)

\bibitem{long2021bayesian}
Long, Z., Zhu, C., Liu, J., Liu, Y.: Bayesian low rank tensor ring for image
  recovery. IEEE Transactions on Image Processing  \textbf{30},  3568--3580
  (2021)

\bibitem{nickel2011three}
Nickel, M., Tresp, V., Kriegel, H.P.: A three-way model for collective learning
  on multi-relational data. In: Proceedings of the 28th International
  Conference on International Conference on Machine Learning. pp. 809--816
  (2011)

\bibitem{oh2018scalable}
Oh, S., Park, N., Lee, S., Kang, U.: Scalable tucker factorization for sparse
  tensors-algorithms and discoveries. In: 2018 IEEE 34th International
  Conference on Data Engineering (ICDE). pp. 1120--1131. IEEE (2018)

\bibitem{oseledets2011tensor}
Oseledets, I.V.: Tensor-train decomposition. SIAM Journal on Scientific
  Computing  \textbf{33}(5),  2295--2317 (2011)

\bibitem{polson2013bayesian}
Polson, N.G., Scott, J.G., Windle, J.: Bayesian inference for logistic models
  using p{\'o}lya--gamma latent variables. Journal of the American statistical
  Association  \textbf{108}(504),  1339--1349 (2013)

\bibitem{rai2015scalable}
Rai, P., Hu, C., Harding, M., Carin, L.: Scalable probabilistic tensor
  factorization for binary and count data. In: IJCAI. pp. 3770--3776 (2015)

\bibitem{rai2014scalable}
Rai, P., Wang, Y., Guo, S., Chen, G., Dunson, D., Carin, L.: Scalable bayesian
  low-rank decomposition of incomplete multiway tensors. In: International
  Conference on Machine Learning. pp. 1800--1808. PMLR (2014)

\bibitem{schein2016bayesian}
Schein, A., Zhou, M., Blei, D., Wallach, H.: Bayesian poisson tucker
  decomposition for learning the structure of international relations. In:
  International Conference on Machine Learning. pp. 2810--2819. PMLR (2016)

\bibitem{tao2020bayesian}
Tao, Z., Zhao, Q.: Bayesian tensor ring decomposition for low rank tensor
  completion. In: International Workshop on Tensor Network Representations in
  Machine Learning, IJCAI (2020)

\bibitem{tao2021bayesian}
Tao, Z., Zhao, X., Tanaka, T., Zhao, Q.: Bayesian latent factor model for
  higher-order data. In: Asian Conference on Machine Learning. pp. 1285--1300.
  PMLR (2021)

\bibitem{tucker1966some}
Tucker, L.R.: Some mathematical notes on three-mode factor analysis.
  Psychometrika  \textbf{31}(3),  279--311 (1966)

\bibitem{wang2017efficient}
Wang, W., Aggarwal, V., Aeron, S.: Efficient low rank tensor ring completion.
  In: Proceedings of the IEEE International Conference on Computer Vision. pp.
  5697--5705 (2017)

\bibitem{wang2018wide}
Wang, W., Sun, Y., Eriksson, B., Wang, W., Aggarwal, V.: Wide compression:
  Tensor ring nets. In: Proceedings of the IEEE Conference on Computer Vision
  and Pattern Recognition. pp. 9329--9338 (2018)

\bibitem{xu2021probabilistic}
Xu, L., Cheng, L., Wong, N., Wu, Y.C.: Probabilistic tensor train decomposition
  with automatic rank determination from noisy data. In: 2021 IEEE Statistical
  Signal Processing Workshop (SSP). pp. 461--465. IEEE (2021)

\bibitem{yuan2019tensor}
Yuan, L., Li, C., Mandic, D., Cao, J., Zhao, Q.: Tensor ring decomposition with
  rank minimization on latent space: An efficient approach for tensor
  completion. In: Proceedings of the AAAI conference on artificial
  intelligence. vol.~33, pp. 9151--9158 (2019)

\bibitem{yuan2017completion}
Yuan, L., Zhao, Q., Cao, J.: Completion of high order tensor data with missing
  entries via tensor-train decomposition. In: Neural Information Processing:
  24th International Conference, ICONIP 2017, Guangzhou, China, November 14-18,
  2017, Proceedings, Part I. pp. 222--229. Springer (2017)

\bibitem{zhao2015bayesian}
Zhao, Q., Zhang, L., Cichocki, A.: Bayesian cp factorization of incomplete
  tensors with automatic rank determination. IEEE transactions on pattern
  analysis and machine intelligence  \textbf{37}(9),  1751--1763 (2015)

\bibitem{zhao2015bayesiantucker}
Zhao, Q., Zhang, L., Cichocki, A.: Bayesian sparse tucker models for dimension
  reduction and tensor completion. arXiv preprint arXiv:1505.02343  (2015)

\bibitem{zhao2016tensor}
Zhao, Q., Zhou, G., Xie, S., Zhang, L., Cichocki, A.: Tensor ring
  decomposition. arXiv preprint arXiv:1606.05535  (2016)

\bibitem{zhe2015scalable}
Zhe, S., Xu, Z., Chu, X., Qi, Y., Park, Y.: Scalable nonparametric multiway
  data analysis. In: Artificial Intelligence and Statistics. pp. 1125--1134.
  PMLR (2015)

\end{thebibliography}




% LNCS Homepage, \url{http://www.springer.com/lncs}. Last accessed 4
% \end{thebibliography}
\end{document}